\documentclass[sigconf]{acmart}

\usepackage{multirow}
\usepackage{colortbl}

\AtBeginDocument{%
  }

\setcopyright{acmlicensed}
\copyrightyear{2025}
\acmYear{2025}
\acmDOI{XXXXXXX.XXXXXXX}
\acmConference[Conference acronym 'XX]{Make sure to enter the correct
  conference title from your rights confirmation email}{ October 27--31,
  2025}{Woodstock, Ireland}

\acmISBN{979-8-4007-2035-2/2025/10}

\usepackage{subcaption}

\copyrightyear{2025}
\acmYear{2025}
\setcopyright{cc}
\setcctype{by}
\acmConference[MM '25]{Proceedings of the 33rd ACM International Conference on Multimedia}{October 27--31, 2025}{Dublin, Ireland}
\acmBooktitle{Proceedings of the 33rd ACM International Conference on Multimedia (MM '25), October 27--31, 2025, Dublin, Ireland}\acmDOI{10.1145/3746027.3754557}
\acmISBN{979-8-4007-2035-2/2025/10}

\begin{document}

\title[Unlocking Financial Insights: An advanced Multimodal Summarization with Multimodal Output Framework \\ for Financial Advisory Videos]{Unlocking Financial Insights: An advanced Multimodal Summarization with Multimodal Output Framework for Financial Advisory Videos}

\author{Sarmistha Das}
\authornote{The first two authors contributed equally to this work and are jointly the first authors.}
\email{sarmistha_2221cs21@iitp.ac.in}
\orcid{0000-0001-6608-0400}
\affiliation{
  \institution{Indian Institute of Technology Patna}
  \city{Patna}
  \country{India}
}

\author{R. E. Zera Marveen Lyngkhoi }
\authornotemark[1]
\email{zeramarveenlyngkhoi@gmail.com}
\affiliation{
  \institution{Indian Institute of Technology Patna}
  \city{Patna}
  \country{India}
}

\author{Sriparna Saha}
\email{sriparna.saha@gmail.com}
\orcid{0000-0001-5458-9381}
\affiliation{%
  \institution{Indian Institute of Technology Patna}
  \city{Patna}
  \country{India}
}

\author{Alka Maurya}
\affiliation{%
  \institution{CRISIL LTD}
  \city{Mumbai}
  \country{India}}

\renewcommand{\shortauthors}{Sarmistha Das, R. E. Zera Marveen Lyngkhoi, Sriparna Saha, and Alka Maurya}
\begin{abstract}
The dynamic propagation of social media has broadened the reach of financial advisory content through podcast videos, yet extracting insights from lengthy, multimodal segments (30–40 minutes) remains challenging. We introduce \textit{FASTER} (Financial Advisory Summariser with Textual Embedded Relevant images), a modular framework that tackles three key challenges: (1) extracting modality-specific features, (2) producing optimized, concise summaries, and (3) aligning visual keyframes with associated textual points. \textit{FASTER} employs \textit{\underline{B}LIP-2} for semantic visual descriptions, \textit{\underline{O}CR} for textual patterns, and Whisper-based transcription with \textit{\underline{S}peaker} diarization as BOS features. A modified Direct Preference Optimization (DPO)-based loss function, equipped with BOS-specific fact-checking, ensures precision, relevance, and factual consistency against the human-aligned summary. A ranker-based retrieval mechanism further aligns keyframes with summarized content, enhancing interpretability and cross-modal coherence. 
To acknowledge data resource scarcity, we introduce {\textit{Fin-APT}}, a dataset comprising 470 publicly accessible financial advisory pep-talk videos for robust multimodal research. Comprehensive cross-domain experiments confirm \textit{FASTER}’s strong performance, robustness, and generalizability when compared to Large Language Models (LLMs) and Vision-Language Models (VLMs). By establishing a new standard for multimodal summarization, \textit{FASTER} makes financial advisory content more accessible and actionable, thereby opening new avenues for research\footnote{The dataset and code are available at: \url{https://github.com/sarmistha-D/FASTER}}.
\end{abstract}







\maketitle
\section{Introduction}

In the FinTech-3.5 era, financial advisory podcasts have emerged as a vital channel for delivering wealth management strategies to teenage and young adult demographics \cite{podcast}. 
In the United States, 46\% of Gen Xers fear inadequate retirement reserves; research indicates that professional guidance on retirement planning, asset allocation, and risk mitigation drives an annual 7.5\% increase in household savings and demonstrably lowers financial stress. Leading financial podcast platforms (Spotify, Apple Music, YouTube) deliver concise (~40-minutes) episodes featuring experts like Warren Buffett, Dave Ramsey, and Robert Kiyosaki. These broadcasts integrate real-time market data, tax strategies, mature investment perspectives, and cutting-edge technology to elevate financial literacy, bolster entrepreneurial acumen, and foster sustainable wealth creation.
\begin{figure*}
    \centering
    \includegraphics[scale=0.33]{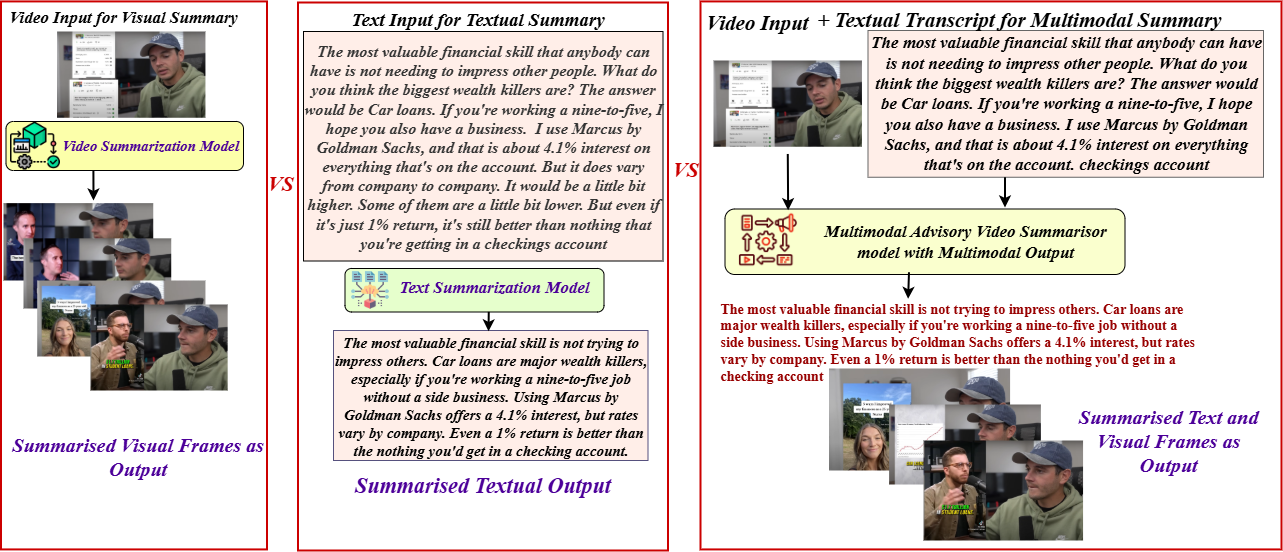}
    \caption{Comparison of Summarisation Approaches: Video-Based, Text-Based, and Multimodal Methods with Multimodal Output Generation}
    \label{model_example}
\end{figure*}
Occasionally, lengthy video content can often become monotonous for young entrepreneurs with limited financial exposure, making it challenging for them to grasp financial wisdom. Summarizing financial podcast videos enables users to quickly capture key insights on marketing strategies, emerging trends, and savings frameworks without needing to watch the entire content, as illustrated in Figure \ref{model_example}. However, summarizing multimodal content, non-linear video elements (e.g., flashbacks and discussion cuts), and critical financial data (e.g., stock prices) embedded in textual and visual formats plays a pivotal role in delivering concise, context-aware briefings tailored to user-specific needs. Often multi-modal output helps to select 
 the most relevant keyframe while condensing textual content. LLMs have redefined NLP, showcasing potential for artificial general intelligence in finance \cite{brown2020language,touvron2023llama}. Models like BloombergGPT \cite{wu2023bloomberggpt}, FinGPT \cite{liu2023fingpt}, and FinMA \cite{xie2023pixiu} enhance financial decision-making by summarizing complex data but are constrained by unimodal design and restricted access to large-parameter architectures.\\ 
 \textbf{\textit{Motivation:}} Condensing financial videos with LLM models is complex due to the integration of multimodal data, particularly in scenarios like financial podcasts where static visuals provide minimal context \cite{xu2015jointly}. Current models fail to utilize visual cues effectively, missing key insights for accurate summarization. Furthermore, there is a lack of accessible datasets featuring expert-driven financial advisory content. To address these challenges, we propose \textit{FASTER}, a modular framework that combines visual (BLIP-2, OCR), acoustic, and textual features. Utilizing a curriculum learning embedded modified Direct Preference Optimization (DPO) \cite{rafailov2023direct} and a ranking algorithm, \textit{FASTER} ensures coherent multimodal summaries by aligning critical textual information with relevant visual keyframes, enhancing both accuracy and context.

\noindent\textbf{Contributions:} The major contributions introduced to our summary generation model are given below:\\
    1) Introducing \textit{Fin-APT}, a pioneering multimodal dataset of 470 meticulously annotated financial advisory videos, categorized by domains and voice tones. As the first of its kind, it bridges critical gaps in multimodal financial research, enabling advanced discourse analysis and innovation in communication modeling.\\
    2) We introduce \textit{FASTER}, a modular summarization model leveraging visual, acoustic, and textual modalities via BLIP-2, OCR, and speaker identification as BOS features. Employing an enhanced DPO framework with fact score, \textit{FASTER} produces precise, high-quality multimodal summaries, seamlessly aligning textual insights with key visual frames.\\

\begin{figure}[h]    
\centering
    \includegraphics[scale = 0.42]{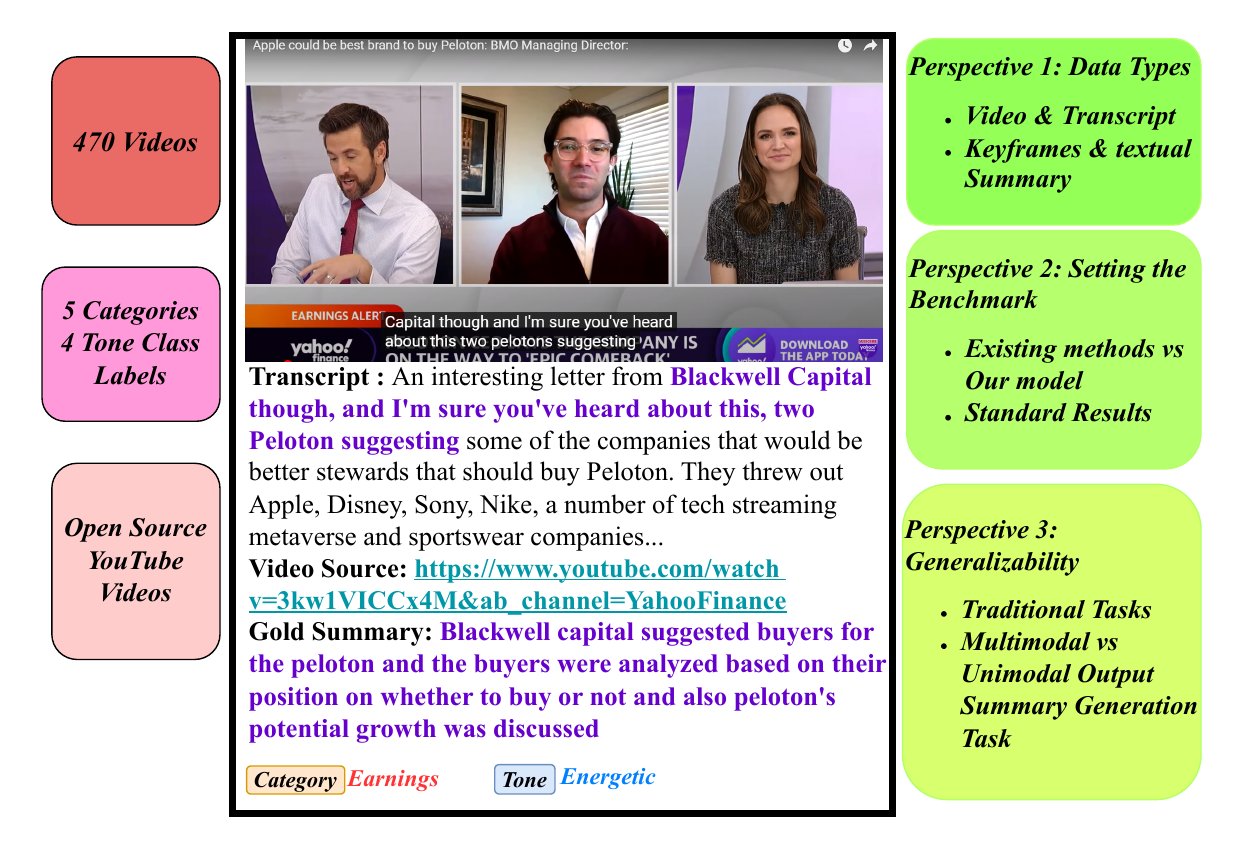}
    \caption{The construction of the proposed Fin-APT dataset is guided by both research objectives and practical application requirements.} 
    \label{Fin-APT}
\end{figure}

\section{Background Information}\label{related studies}
\textbf{Finance in the LLM Era:} Social media platforms like Twitter have amplified the visibility of sustainable startups, increasing demand for financial advisory services \cite{franco2021innovation,passali2021towards,abdaljalil2021exploration}. Advanced domain-specific models like FinBERT \cite{yang2020finbert} and FinConnect \cite{ceadar_2023}, have revolutionized financial text analysis, While tools like InvestLLM \cite{yang2023investlm}, leveraging instruction tuning, and AI agents like FinRobot \cite{yang2024finrobot} demonstrate AI's role in executing complex financial strategies.

\noindent\textbf{Financial Summarization: }Summarizing actionable insights is essential in finance \cite{gokhan2021extractive}. Multi-document summarization, established in 2015 \cite{zhong2015query}, enabled tailored models like Pan et al.'s trend forecasting \cite{pan2021roles} and statistical approaches for marketing strategies. Shared tasks like FNS \cite{zavitsanos2023financial} and FinLLM \cite{xie2024finnlp} further illustrate LLMs' potential. Yet, summarizing non-textual financial content, such as videos and audio, remains an unmet need.

\noindent\textbf{Multimodality in Finance: }Multimodal approaches integrate text, audio, numerical, and visual data to enhance predictive accuracy \cite{li2020maec,sawhney2020multimodal}. Evolving from clustering-based methods \cite{wang2011integrating} to unsupervised deep learning \cite{zhong2015query}, these techniques have shown efficacy in risk forecasting and market volatility prediction \cite{cheng2022financial}. Visual data also boosts engagement, with pictorial summaries increasing satisfaction by 12.4\% \cite{zhu2018msmo}. However, despite advancements in video summarization and visual language models \cite{xu2015show,tong2022videomae}, financial multimodal content remains underexplored.
\\
\noindent\textbf{\textit{Our Research Position:}} The increasing reliance on LLMs and multimodal frameworks highlights the need for factually aligned, privacy-focused summaries in finance. Multimodal summarization with multimodal output is largely unexplored in this domain \cite{jangra2023survey,tang2023tldw}. Addressing this gap, we aim to deliver optimized, fact-checked multimodal summaries using DPO and robust verification mechanisms to ensure ethical and reliable solutions.

\begin{table*}[h]
\caption{Comparison of Existing Datasets in Finance Realm}\label{data_comparison}
\centering
\scalebox{0.70}{

\begin{tabular}{l|l|lll|ll|l|r}
\hline
Modality                                                  & Dataset Name                                            & \multicolumn{3}{l|}{Input}                                           & \multicolumn{2}{l|}{Output}             & Task          & \multicolumn{1}{l}{Sample Count} \\ \cline{3-7}
      &                                                               & \multicolumn{1}{l|}{Text} & \multicolumn{1}{l|}{Audio} & Image/Video & \multicolumn{1}{l|}{Text} & Image/Video &       & \multicolumn{1}{l}{}   \\ \hline
Unimodal                                                                   & Trading Dataset \cite{fabbri2018dow}                                                         & \multicolumn{1}{l|}{$\checkmark$}    & \multicolumn{1}{l|}{$\times$ }     & $\times$           & \multicolumn{1}{l|}{$\checkmark$}    & $\times$            & Tading Task                    & 3285                                               \\ \hline
Audio Calls                                                                & Volatility Prediction \cite{yang2020html}                                                   & \multicolumn{1}{l|}{$\checkmark$}    & \multicolumn{1}{l|}{$\checkmark$}     & $\times$             & \multicolumn{1}{l|}{$\checkmark$}    & $\times$     & Stock prediction               & 576                                                \\ \hline
Audio Calls                                                                & MAECS\&P 1500 \cite{yang2020html}                                                           & \multicolumn{1}{l|}{$\checkmark$}    & \multicolumn{1}{l|}{$\checkmark$}     & $\times$             & \multicolumn{1}{l|}{$\checkmark$}    & $\times$             &  Volatility Prediction                          & 576                                                \\ \hline
\begin{tabular}[c]{@{}l@{}}Audio Calls \&\\  Transcript pairs\end{tabular} & \begin{tabular}[c]{@{}l@{}}Multitasking Risk \\ Forecasting \cite{sawhney2020multimodal}\end{tabular} & \multicolumn{1}{l|}{$\checkmark$}    & \multicolumn{1}{l|}{$\checkmark$}     & $\times$            & \multicolumn{1}{l|}{$\checkmark$}    & $\times$             &  Risk Forecasting          & 9383                                               \\ \hline
Text  & X-FinCORP \cite{das2023let}                                                               & \multicolumn{1}{l|}{$\checkmark$}    & \multicolumn{1}{l|}{$\times$  }     & $\times$            & \multicolumn{1}{l|}{$\checkmark$}    & $\times$             & Complaint Mining               & 6282                                               \\ \hline

Image & TERED \cite{das2023find}                                                                    & \multicolumn{1}{l|}{$\times$  }    & \multicolumn{1}{l|}{$\times$  }     & $\checkmark$           & \multicolumn{1}{l|}{$\checkmark$}    & $\checkmark$           & Financial table Detection      & 10,044                                             \\ \hline
Text  & EDTSum \cite{xie2024finnlp}                                                                  & \multicolumn{1}{l|}{$\checkmark$}    & \multicolumn{1}{l|}{$\times$  }     & $\times$            & \multicolumn{1}{l|}{$\checkmark$}    & $\times$             & Summarisation                  & \multicolumn{1}{l}{}  10,000                            \\ \hline
Video  & \textit{Fin-APT (Proposed)}                                                                 & \multicolumn{1}{l|}{{\color{green}{$\checkmark$}}}    & \multicolumn{1}{l|}{{\color{green}{$\checkmark$}}}     & {\color{green}{$\checkmark$}}          & \multicolumn{1}{l|}{{\color{green}{$\checkmark$}}}    & {\color{green}{$\checkmark$}}          &  Summarisation                              & \multicolumn{1}{l}{}   470   \\ \hline

\end{tabular}}
\end{table*}

\section{Development of the Resource}
YouTube, with its 2.9 billion global users, has emerged as a pivotal platform for individuals seeking financial guidance, offering extensive content on wealth accumulation, investment strategies, money management, and passive income generation. Our initial objective was to identify key domains where users predominantly seek credible financial advice and quantify these trends through systematic data collection. A comprehensive review of existing datasets revealed a critical gap for a specialized multimodal resource focused on summarizing financial advisory videos. Table \ref{data_comparison} underscores this gap by comparing recent datasets. To address this need, we curated the \textit{Fin-APT} multimodal dataset \footnote{\textbf{Copyright Disclaimer: The dataset includes video links copyrighted by YouTube, intended strictly for research purposes.}}, inspired by the intricate and diverse landscape of online financial advice.\\
\textit{\textbf{Dataset Collection:}}
The \textit{Fin-APT} dataset consists of 470 high-definition videos featuring expert interviews on current market trends and actionable insights for individual financial strategies. Initially, 700 videos across diverse financial topics were collected, but through rigorous filtering, the dataset was refined to 470 English-only videos, excluding multilingual content. Data acquisition was conducted primarily using the YouTube API. The dataset covers financial advisory videos from five domains: Business (75), Finance (217), Investment (113), Economics (35), and Marketing (30). Additionally, the dataset includes videos categorized by four distinct voice tones: Informative (149), Neutral (60), Energetic (211), and Cautious (50).\\ 
\textit{\textbf{Dataset Specifications:}}
Each video in the \textit{Fin-APT} dataset was transcribed using OpenAI's robust \textit{Whisper} model \cite{radford2023robust}, ensuring precise transcripts with word-level timestamps. Post-transcription, the quality was enhanced by refining unclear words and removing disruptions such as stammering or filler symbols. The videos in the dataset are capped at a maximum duration of 2400 seconds. Figure \ref{Fin-APT} presents a comprehensive overview of the dataset's key characteristics.\\
\textit{\textbf{Corpus Authenticity Measurement}}
To establish a gold standard summary, a focused evaluation was conducted emphasizing relevance and coherence, assessed through adequacy and fluency metrics. The evaluation utilized E-FAIR (Evaluation of Fluency, Adequacy, Informativeness, and Persuasiveness Ratings) scores on a 0–5 scale: 0–1 (Inferior), 1–2 (Poor), 2–3 (Moderate), 3–4 (Good), and 4–5 (Excellent). Additionally, visual cues (highlighted on-screen information) and audio elements (speaker details) were analyzed to ensure factual accuracy and information retention.
\subsection{Corpus Annotation}
The annotation of the entire corpus was conducted in the following three phases: 
\begin{itemize}
    \item \textbf{Annotator Selection}: The gold standard summarization was developed with input from expert annotators in \textit{Category A}, comprising three specialists: two industry experts from S\&P Global and an economics professor from a leading technology institute. These experts are regular Spotify users, actively engaged with financial podcasts on topics such as investment strategies and tax optimization, typically ranging from 35 to 45 minutes in duration. Additionally, a two-phase annotation competition was conducted to recruit junior annotators from finance and science backgrounds. Out of 20 participants, a rigorous evaluation and training process culminated in the selection of two winners, who were designated as the final \textit{Category B} annotators.
    \item \textbf{Scrutinization Phase:} Experts initially created summaries for 60 randomly selected video samples, refining them through mutual discussion and quality assessment to establish a gold-standard reference set. The annotator selection process involved two subphases: (a) \textbf{Preliminary Selection}, where annotators were provided with 20 reference samples and detailed annotation guidelines, and (b) \textbf{Final Selection}, where annotators independently annotated 30 samples. After rigorous quality evaluation, 7 annotators advanced to the final phase, culminating in the selection of 2 top-performing annotators. 
    \item \textbf{Training Phase:} During the preliminary selection phase, annotators adhered to the prescribed annotation guidelines independently, with any queries addressed by \textit{Category A} experts. Annotators were instructed to preserve visual and speaker information in their summary labels. In the final selection phase, summaries were evaluated by \textit{Category A} annotators using E-FAIR and visual-audio fact scores. The gold standard was established by selecting summaries with the highest ratings. Annotators were compensated at \$0.5 per sample.

\end{itemize}

\section{Methodology}

\begin{figure*}[!hbt]
    \centering
    \includegraphics[scale = 0.39]{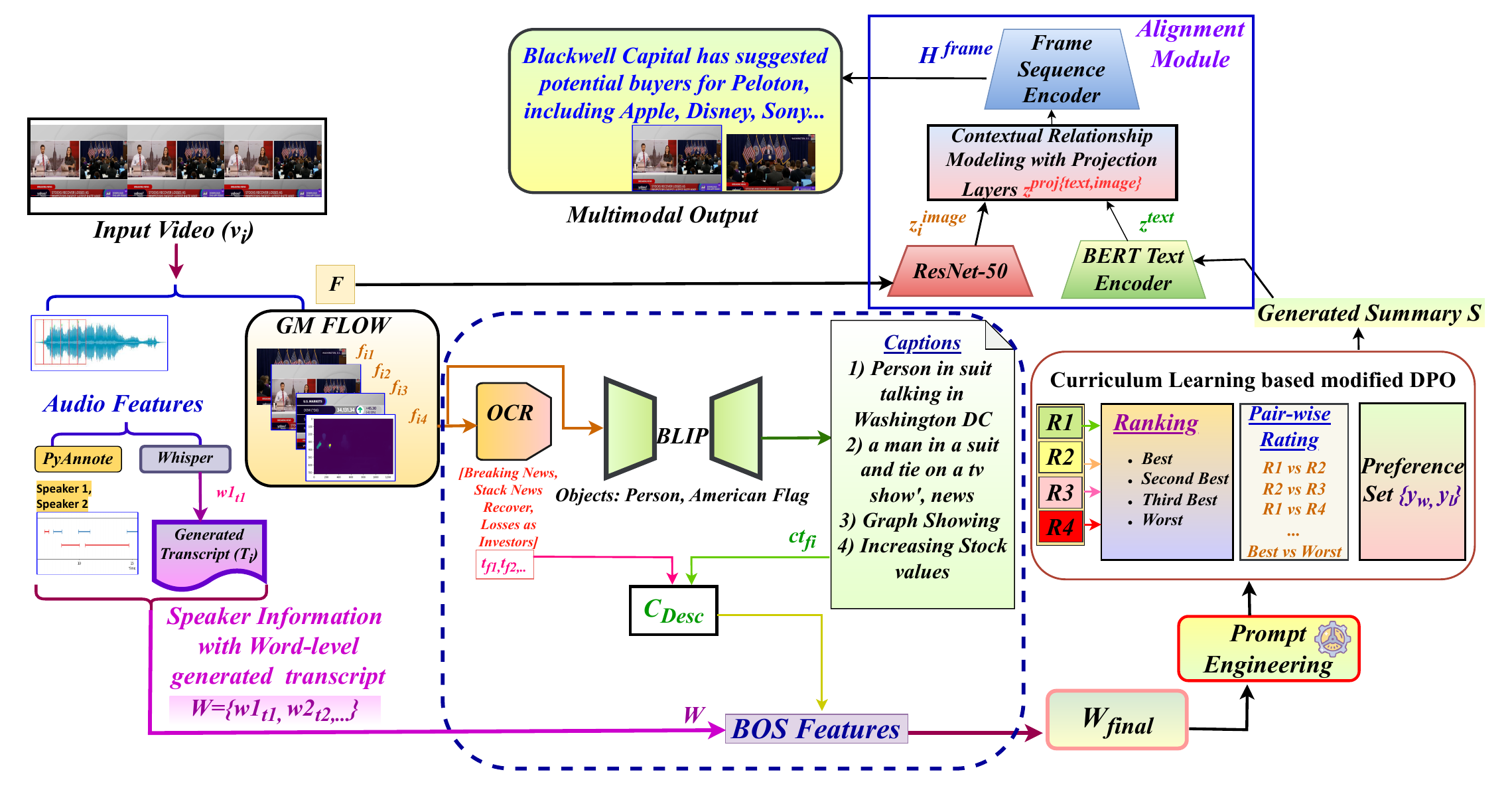}
    \caption{Architecture of the proposed\textit{ FASTER } based Financial multi-modal summary generation with multi-modal output}\label{archi}
\end{figure*}

\textbf{\textit{Problem Defination}}
Given a set of financial advisory videos \( V = \{v_1, v_2, \ldots, v_n\} \), where each video \( v_i \) consists of visual frames \( F_i = \{f_{i1}, f_{i2}, \ldots, f_{im}\} \) and corresponding transcript text \( T_i \) from the audio stream, the objective is to generate a coherent and brief multimodal summary \( S_i = (S_i^{text}, S_i^{images}) \) for each \( v_i \), where \( S_i^{text} = \{s_{i1}, s_{i2}, \ldots, s_{il}\} \) represents the textual summary limited by the max length of $l$ and \( S_i^{images} = \{I_{i1}, I_{i2}, \ldots, I_{ik}\} \) represents the subset of most relevant images from the input video frames $F_i$. Figure \ref{archi} depicts the entire workflow of the model.

  \begin{figure}[!hbt]    
    \centering
    \includegraphics[scale = 0.32]{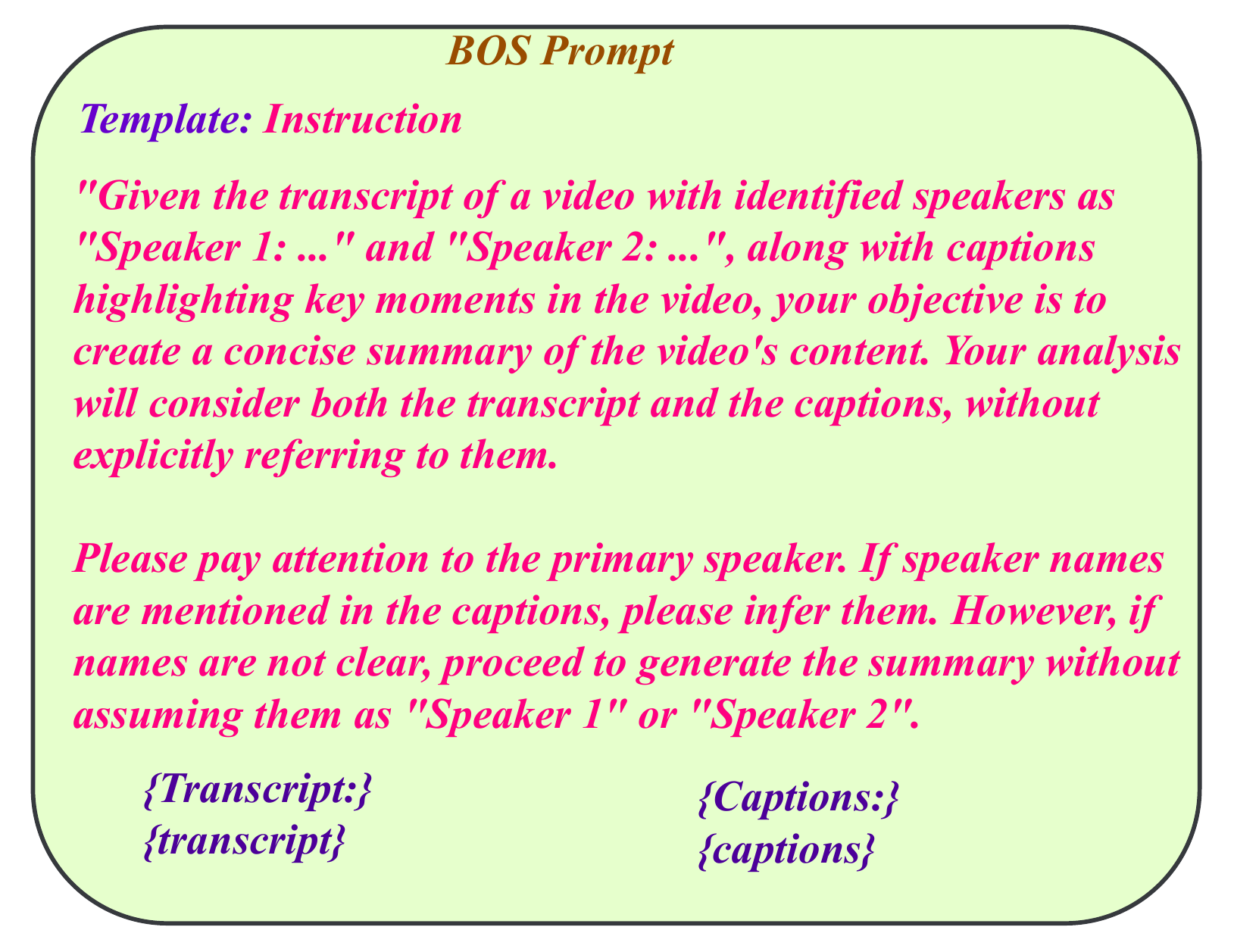}
    \caption{The construction of the proposed BOS(BLIP-2 + OCR+ Speaker Information) features prompt engineering.} 
    \label{BOS-Prompt}
\end{figure}

\begin{itemize}
       \item \textbf{Crafting the BOS (BLIP-2+OCR+Speaker Information) features:}
    The Visual Context Generation Module extracts relevant visual features from video sequences by first applying the GM-Flow \cite{xu2022gmflow} technique to identify key frames based on significant optical flow changes, prioritizing frames with the highest pixel velocity to capture dynamic segments, forming the subset $F_{s}=\{f_i\}$. Each selected frame undergoes OCR to extract textual content, yielding $T_{OCR}=\{t_{f1}, t_{f2},...\}$. Subsequently, BLIP-2-7B \cite{li2023blip2}, a vision-language model, generates visual captions $ct_{fi}$, which are combined with OCR text $t_{fi}$ to produce comprehensive frame descriptions $C_{Desc}=\{c_{f1}, c_{f2},...\}$. The Audio Processing Module enhances understanding of auditory content via Pyannote \cite{Bredin2020} for speaker diarization, segregating audio by speaker identity with timestamps to distinguish speaker-specific content. The transcribed audio is processed using Whisper, generating word-level transcriptions with timestamps, $W=\{w1_{t1}, w2_{t2},...\}$. The diarization results are integrated to attribute each transcribed segment to its corresponding speaker, producing the final enriched transcription $W_{final}$ that encapsulates the audio’s semantic and contextual details, thus presents a comprehensive view of who said what and when encapsulating both the content and the context of the auditory narrative. Figure~\ref{BOS-Prompt} illustrates the prompt engineering used to extract BOS features.
  
\end{itemize}
\subsection{Summary Generation Module}
To generate high-quality summaries, we utilised extracted BOS features (\( W_{\text{final}} \)) and incorporated them into a prompt and fed to Gemma2-9b \cite{team2024gemma} model to generate summaries. The model was fine-tuned using instruction-based learning, followed by iterative training to further improve its performance. Drawing inspiration from curriculum learning, we modified the Direct Preference Optimization (DPO) framework to enhance the alignment and factual coherence of the model's outputs. This process is conducted in three substeps:\\
\textbf{\textit{Step1 The Ranking Step:}} Initially, fine-tuned Gemma2-9b model \cite{team2024gemma} generated four candidate summaries (\( R_1, R_2, R_3, R_4 \)) for each BOS-featured inputs.
To assess the quality of the summaries, we employed GPT-4o \cite{openai_gpt4}, which ranked the responses according to their quality, coherence, and how well they aligned with human-annotated gold-standard summaries. 
Let us assume, the responses were ranked in the following order: \( R_1 \) was considered the best, followed by \( R_2 \) as the second-best, \( R_3 \) as the third-best, and \( R_4 \) as the worst response. Subsequently, to validate GPT-4o's ranking quality, human evaluation confirmed the accuracy of the ranking. 
\\
\textbf{\textit{Step2 Training of Pairwise Preferences:}}
To enhance the model's ability to identify high-quality summaries, pairwise preference training was applied using a curriculum learning strategy. The process began with the comparison of \( R_1 \) (best) vs. \( R_4 \) (worst), where \( R_1 \) was labeled as the chosen response (\( y_w \)) and \( R_4 \) as the rejected response (\( y_l \)), helping the model distinguish between the highest and lowest-quality summaries. In the next iteration, \( R_1 \) was compared with \( R_3 \) (third-best), followed by a final comparison between \( R_1 \) and \( R_2 \) (second-best), gradually advancing to subtler distinctions.
\\
\textbf{\textit{Step3 Training with Loss Function:}} To refine preferences by comparing the probabilities of preferred and less preferred responses, we utilised the DPO loss function, where a LLM is used to select the preferred output \( y_w \) and less preferred output \( y_l \), can be expressed as:
\textbf{
\begin{multline*}
L_{\text{DPO}}(\pi_\theta; \pi_{\text{LLM}}) = \\
- \mathbb{E}_{(x, y_w, y_l) \sim D_{\text{LLM}}} \left[ \log \sigma \left( \beta \log \frac{\pi_\theta(y_w \mid x)}{\pi_{\text{LLM}}(y_w \mid x)} \right. \right. \\
\left. \left. - \beta \log \frac{\pi_\theta(y_l \mid x)}{\pi_{\text{LLM}}(y_l \mid x)} \right) \right]
\end{multline*}
}
In this framework, \( \pi_\theta(y_w \mid x) \) and \( \pi_\theta(y_l \mid x) \) denote the probabilities assigned by the model's policy \( \pi_\theta \) to the preferred and less preferred responses, while \( \pi_{\text{LLM}}(y_w \mid x) \) and \( \pi_{\text{LLM}}(y_l \mid x) \) represent the corresponding probabilities from the reference large language model (LLM). The parameter \( \beta \) controls the sensitivity of the preference distinction, scaling the sharpness of the comparison. The sigmoid function \( \sigma(z) = \frac{1}{1 + e^{-z}} \) maps the log-probability difference to a range between 0 and 1, quantifying the likelihood of the preferred response (\( y_w \)) being favored over the less preferred one (\( y_l \)).

We use curriculum learning with a modified-DPO approach to enable the model to progress from simple distinctions to more complex comparisons. The loss function for iteration \( i+1 \) in modified-DPO is:
\textbf{
\begin{multline*}
L_{\text{Modified\_DPO}}(\pi_\theta^{i+1}; \pi_\theta^i) = \\
- \mathbb{E}_{(x, y_w^{i+1}, y_l^{i+1}) \sim D} \left[ \log \sigma \left( \beta \log \frac{\pi_\theta^{i+1}(y_w \mid x)}{\pi_\theta^i(y_w \mid x)} \right. \right. \\
\left. \left. - \beta \log \frac{\pi_\theta^{i+1}(y_l \mid x)}{\pi_\theta^i(y_l \mid x)} \right) \right]
\end{multline*}
}
Where, \( \pi_\theta^i \) is the reference model from iteration \( i \), while \( \pi_\theta^{i+1} \) is the policy model being trained in iteration \( i+1 \). During each iteration, chosen (\( y_w^{i+1} \)) and rejected (\( y_l^{i+1} \)) pairs are selected based on the curriculum strategy, which progressively exposes the model to more complex tasks.
\subsection{Ranking Base Multimodal Output}
To generate aligned multimodal outputs, a ranking-based alignment approach prioritized video frames \( F = \{f_1, f_2, \dots, f_N\} \) most relevant to the curriculum learning based modified-DPO generated text summary \( S \). Visual features \( z_i^{\text{image}} \) were extracted from each frame \( f_i \) using ResNet-50 \cite{he2016deep} and projected into a 512-dimensional space \( Z^{\text{proj-image}} \). The text summary was encoded using BERT \cite{devlin-etal-2019-bert}, producing a 768-dimensional embedding \( z^{\text{text}} \), which was also projected into the same 512-dimensional space \( z^{\text{proj-text}} \). Contextual relationships among frames were modeled using a transformer encoder, generating context-aware embeddings \( H^{\text{frame}} = \{h_1^{\text{frame}}, h_2^{\text{frame}}, \dots, h_N^{\text{frame}}\} \). An attention mechanism aligned the text embedding \( z^{\text{proj-text}} \) (query) with the frame embeddings \( H^{\text{frame}} \) (keys and values), computing attention weights \( A = \text{softmax} \left( \frac{QK^\top}{\sqrt{d}} \right) \), where \( Q \), \( K \), and \( V \) are derived from \( z^{\text{proj-text}} \) and \( H^{\text{frame}} \). A scoring layer calculated relevance scores \( s_i = \sigma(W_s h_i^{\text{frame}} + b_s) \) for each frame, and the top \( k \) frames with the highest scores were selected. Training optimized a loss function \( \mathcal{L} = \mathcal{L}_{\text{BCE}} + \lambda \mathcal{L}_{\text{diversity}} \), where \( \mathcal{L}_{\text{BCE}} \) aligns scores with ground-truth relevance and \( \mathcal{L}_{\text{diversity}} = \| A A^\top - I \|_F^2 \) encourages diversity in frame selection. During inference, frames were ranked by \( s_i \), and the top \( k \) frames were chosen, ensuring effective alignment with the text summary.
\section{Experiments \& Resultant Discussion}

\begin{table*}[]
\caption{The variance in performance metrics among distinct baseline LLM models with the proposed \textit{FASTER} model configuration is presented below. In this context, R-1, and R-2 signify ROUGE-1, ROUGE-2 score.  B-1, B-1, B-2, B-3, B-4 indicate BLEU-1, BLEU-2, BLEU-3, BLEU-4 and BS denotes BERTSCORE. Values in bold signify the highest scores achieved. The † denotes statistically significant findings ($p < 0.05 $ at 5\% significance level)}\label{comp}
\centering
\setlength{\tabcolsep}{0.85pt}
\renewcommand{\arraystretch}{1.5}
\scalebox{0.65}{

\begin{tabular}{c|c|ccccccc|ccccccc|ccccccc}
\hline
{\textbf{Modality of the Models}} & {\textbf{Model Names}} & \multicolumn{7}{c|}{\textbf{Base Setting}}  & \multicolumn{7}{c|}{\textbf{BOS Injected}} & \multicolumn{7}{c}{\textbf{BOS+DPO Injected}}     \\ \cline{3-23} 
 && \multicolumn{1}{c|}{\textbf{R-1}} & \multicolumn{1}{c|}{\textbf{R-2}} & \multicolumn{1}{c|}{\textbf{B-1}} & \multicolumn{1}{c|}{\textbf{B-2}} & \multicolumn{1}{c|}{\textbf{B-3}} & \multicolumn{1}{c|}{\textbf{B-4}} & \textbf{BS} & \multicolumn{1}{c|}{\textbf{R-1}}    & \multicolumn{1}{c|}{\textbf{R-2}}   & \multicolumn{1}{c|}{\textbf{B-1}}  & \multicolumn{1}{c|}{\textbf{B-2}}   & \multicolumn{1}{c|}{\textbf{B-3}}   & \multicolumn{1}{c|}{\textbf{B-4}}   & \textbf{BS}    & \multicolumn{1}{c|}{\textbf{R-1}}   & \multicolumn{1}{c|}{\textbf{R-2}}   & \multicolumn{1}{c|}{\textbf{B-1}}   & \multicolumn{1}{c|}{\textbf{B-2}}   & \multicolumn{1}{c|}{\textbf{B-3}}   & \multicolumn{1}{c|}{\textbf{B-4}}   & \textbf{BS}    \\ \hline
{\textbf{Unimodal}}& \textbf{BART} & \multicolumn{1}{c|}{14.1} & \multicolumn{1}{c|}{5.23} & \multicolumn{1}{c|}{8.73} & \multicolumn{1}{c|}{25.78}& \multicolumn{1}{c|}{22.51}& \multicolumn{1}{c|}{18.81}& 75.91& \multicolumn{1}{c|}{22.23}   & \multicolumn{1}{c|}{07.08}  & \multicolumn{1}{c|}{14.6}  & \multicolumn{1}{c|}{44.16}  & \multicolumn{1}{c|}{37.59}  & \multicolumn{1}{c|}{30.65}  & 79.42  & \multicolumn{1}{c|}{23.03}  & \multicolumn{1}{c|}{09.18}  & \multicolumn{1}{c|}{16.01}  & \multicolumn{1}{c|}{45.96}  & \multicolumn{1}{c|}{38.09}  & \multicolumn{1}{c|}{30.67}  & 82.22  \\ \cline{2-23} 
 & \textbf{FlanT5-Base}& \multicolumn{1}{c|}{17.48}& \multicolumn{1}{c|}{4.78} & \multicolumn{1}{c|}{11.12}& \multicolumn{1}{c|}{25.98}& \multicolumn{1}{c|}{21.91}& \multicolumn{1}{c|}{17.74}& 81.35& \multicolumn{1}{c|}{28.74}   & \multicolumn{1}{c|}{08.73}  & \multicolumn{1}{c|}{18.59} & \multicolumn{1}{c|}{50.38}  & \multicolumn{1}{c|}{42.47}  & \multicolumn{1}{c|}{34.16}  & 84.94  & \multicolumn{1}{c|}{29.77}  & \multicolumn{1}{c|}{08.73}  & \multicolumn{1}{c|}{18.59}  & \multicolumn{1}{c|}{50.38}  & \multicolumn{1}{c|}{42.47}  & \multicolumn{1}{c|}{34.16}  & 84.94  \\ \cline{2-23} 
 & \textbf{GPT3.5-Turbo}& \multicolumn{1}{c|}{37.4} & \multicolumn{1}{c|}{16.12}& \multicolumn{1}{c|}{27.14}& \multicolumn{1}{c|}{57.54}& \multicolumn{1}{c|}{52.33}& \multicolumn{1}{c|}{45.73}& 85.07& \multicolumn{1}{c|}{37.58}   & \multicolumn{1}{c|}{12.46}  & \multicolumn{1}{c|}{23.66} & \multicolumn{1}{c|}{59.60}  & \multicolumn{1}{c|}{53.56}  & \multicolumn{1}{c|}{46.85}  & 87.05  & \multicolumn{1}{c|}{37.58}  & \multicolumn{1}{c|}{12.46}  & \multicolumn{1}{c|}{24.01}  & \multicolumn{1}{c|}{60.66}  & \multicolumn{1}{c|}{54.56}  & \multicolumn{1}{c|}{46.85}  & 88.05  \\ \cline{2-23} 
 & \textbf{LLaMA3-Instruct-1b}      & \multicolumn{1}{c|}{38.32}& \multicolumn{1}{c|}{12.65}& \multicolumn{1}{c|}{23.5} & \multicolumn{1}{c|}{61.27}& \multicolumn{1}{c|}{54.97}& \multicolumn{1}{c|}{47.2} & 86.2& \multicolumn{1}{c|}{39.16}   & \multicolumn{1}{c|}{11.11}  & \multicolumn{1}{c|}{24.27} & \multicolumn{1}{c|}{60.83}  & \multicolumn{1}{c|}{52.33}  & \multicolumn{1}{c|}{47.41}  & 87.92  & \multicolumn{1}{c|}{39.96}  & \multicolumn{1}{c|}{11.81}  & \multicolumn{1}{c|}{23.8}   & \multicolumn{1}{c|}{63.22}  & \multicolumn{1}{c|}{56.81}  & \multicolumn{1}{c|}{49.22}  & 87.72  \\ \cline{2-23} 
 & \textbf{Long T5}      & \multicolumn{1}{c|}{05.71}& \multicolumn{1}{c|}{0.08} & \multicolumn{1}{c|}{4.61} & \multicolumn{1}{c|}{50.54}& \multicolumn{1}{c|}{36.99}& \multicolumn{1}{c|}{23.42}& 75.73& \multicolumn{1}{c|}{22.91}   & \multicolumn{1}{c|}{04.66}  & \multicolumn{1}{c|}{12.96} & \multicolumn{1}{c|}{45.01}  & \multicolumn{1}{c|}{36.88}  & \multicolumn{1}{c|}{30.52}  & 82.05  & \multicolumn{1}{c|}{24.61}  & \multicolumn{1}{c|}{04.54}  & \multicolumn{1}{c|}{13.95}  & \multicolumn{1}{c|}{46.81}  & \multicolumn{1}{c|}{39.26}  & \multicolumn{1}{c|}{32.52}  & 83.97  \\ \cline{2-23} 
 & \textbf{Gemma2-9b}    & \multicolumn{1}{c|}{40.72}& \multicolumn{1}{c|}{16.94}& \multicolumn{1}{c|}{27.86}& \multicolumn{1}{c|}{62.44}& \multicolumn{1}{c|}{55.98}& \multicolumn{1}{c|}{48.39}& 88.76& \multicolumn{1}{c|}{\textbf{45.39\textsuperscript{†}}} & \multicolumn{1}{c|}{\textbf{20.17\textsuperscript{†}}} & \multicolumn{1}{c|}{\textbf{29.9\textsuperscript{†}}} & \multicolumn{1}{c|}{\textbf{67.10\textsuperscript{†}}} & \multicolumn{1}{c|}{\textbf{60.69\textsuperscript{†}}} & \multicolumn{1}{c|}{\textbf{53.26\textsuperscript{†}}} & \textbf{89.51\textsuperscript{†}} & \multicolumn{1}{c|}{\textbf{47.49}{\color{green} \(\uparrow\)}
} & \multicolumn{1}{c|}{\textbf{22.76}{\color{green} \(\uparrow\)}
} & \multicolumn{1}{c|}{\textbf{30.96}{\color{green} \(\uparrow\)}
} & \multicolumn{1}{c|}{\textbf{69.58}{\color{green} \(\uparrow\)}
} & \multicolumn{1}{c|}{\textbf{62.43}{\color{green} \(\uparrow\)}
} & \multicolumn{1}{c|}{\textbf{55.13}{\color{green} \(\uparrow\)}
} & \textbf{89.59}{\color{green} \(\uparrow\)}
 \\ \hline
{Multimodal}      & \textbf{MAF-TAV}  & \multicolumn{1}{c|}{39.04}& \multicolumn{1}{c|}{14.20}& \multicolumn{1}{c|}{25.75}& \multicolumn{1}{c|}{13.30}& \multicolumn{1}{c|}{09.95}& \multicolumn{1}{c|}{05.57}& 82.31& \multicolumn{1}{c|}{-}& \multicolumn{1}{c|}{-}      & \multicolumn{1}{c|}{-}     & \multicolumn{1}{c|}{-}      & \multicolumn{1}{c|}{-}      & \multicolumn{1}{c|}{-}      & -      & \multicolumn{1}{c|}{-}      & \multicolumn{1}{c|}{-}      & \multicolumn{1}{c|}{-}      & \multicolumn{1}{c|}{-}      & \multicolumn{1}{c|}{-}      & \multicolumn{1}{c|}{-}      & -      \\ \cline{2-23} 
 & \textbf{VideoLLaMA2-7b}   & \multicolumn{1}{c|}{24.63}& \multicolumn{1}{c|}{3.46} & \multicolumn{1}{c|}{14.69}& \multicolumn{1}{c|}{66.16}& \multicolumn{1}{c|}{56.13}& \multicolumn{1}{c|}{43.82}& 84.64& \multicolumn{1}{c|}{-}& \multicolumn{1}{c|}{-}      & \multicolumn{1}{c|}{-}     & \multicolumn{1}{c|}{-}      & \multicolumn{1}{c|}{-}      & \multicolumn{1}{c|}{-}      & -      & \multicolumn{1}{c|}{-}      & \multicolumn{1}{c|}{-}      & \multicolumn{1}{c|}{-}      & \multicolumn{1}{c|}{-}      & \multicolumn{1}{c|}{-}      & \multicolumn{1}{c|}{-}      & -      \\ \cline{2-23} 
 & \textbf{VideoLLaVA2-7b}   & \multicolumn{1}{c|}{25.66}& \multicolumn{1}{c|}{4.1}  & \multicolumn{1}{c|}{15.13}& \multicolumn{1}{c|}{69.12}& \multicolumn{1}{c|}{58.66}& \multicolumn{1}{c|}{46.01}& 84.55& \multicolumn{1}{c|}{-}& \multicolumn{1}{c|}{-}      & \multicolumn{1}{c|}{-}     & \multicolumn{1}{c|}{-}      & \multicolumn{1}{c|}{-}      & \multicolumn{1}{c|}{-}      & -      & \multicolumn{1}{c|}{-}      & \multicolumn{1}{c|}{-}      & \multicolumn{1}{c|}{-}      & \multicolumn{1}{c|}{-}      & \multicolumn{1}{c|}{-}      & \multicolumn{1}{c|}{-}      & -      \\ \cline{2-23} 
 & \textbf{GPT-4o}& \multicolumn{1}{c|}{41.29}& \multicolumn{1}{c|}{16.51}& \multicolumn{1}{c|}{26.88}& \multicolumn{1}{c|}{61.43}& \multicolumn{1}{c|}{56.81}& \multicolumn{1}{c|}{50.42}& 89.07& \multicolumn{1}{c|}{-}& \multicolumn{1}{c|}{-}      & \multicolumn{1}{c|}{-}     & \multicolumn{1}{c|}{-}      & \multicolumn{1}{c|}{-}      & \multicolumn{1}{c|}{-}      & -      & \multicolumn{1}{c|}{-}      & \multicolumn{1}{c|}{-}      & \multicolumn{1}{c|}{-}      & \multicolumn{1}{c|}{-}      & \multicolumn{1}{c|}{-}& \multicolumn{1}{c|}{-}    & -\\ \hline
\end{tabular}}
\end{table*}
\begin{table}[]
\caption{Quality Assessment of Ranking alignment for multi-modal image generation.}\label{ranking}
\centering
\scalebox{0.80}{
\begin{tabular}{l|c|c|c}
\hline
\textbf{Method} & \textbf{RMSE} & \textbf{SSIM} & \textbf{F1}  \\ 
\hline
Co-sine Similarity & 02.76  & 0.54 & 0.08  \\
\hline
Trainable Scorer & 0.41 & 0.442 & \textbf{0.32}  \\
\hline

\end{tabular}
}
\end{table}
\begin{table}[hbt!]
\caption{Performance of the proposed \textit{\textit{FASTER}} model with different modalities.  Here, S, T, and V indicate Speaker Information, Transcripts, and Video features, respectively}\label{Modalities}
    \centering
    \scalebox{0.80}{
    \begin{tabular}{lccc}
    \hline
     \textbf{Model}  & \textbf{BERTSCORE} &  \textbf{BLEU} &  \textbf{ROUGE- L}  \\ \hline
     T   & 87.80 & 38.60 & 35.05\\
     V   & 75.12 & 17.8 & 25.1\\
     T + V & 88.01 & 36.31 & 38.70 \\
     V + T + S & \textbf{88.40} & \textbf{39.5} & 4\textbf{0.4}\\ 
     \hline
    \end{tabular}
    }  

\end{table}
\begin{table*}[!t]
\caption{Quality assessment of BOS (BLIP-2+OCR+Speaker Information) features on generated Summary}\label{Bos_comp}
\centering
\scalebox{0.74}{
\begin{tabular}{p{0.45\textwidth}|p{0.44\textwidth}|p{0.4\textwidth}}
\hline
\textbf{Transcript} & \textbf{BOS} & \textbf{Without BOS} \\
\hline
Okay, so tell me this, you write two hours every morning. How does that make you a better investor and portfolio manager? Well, let’s start with this, that this is not the only goal of my life. My goal for life is also to be a good... When you write something, it forces you to see the broken links in your thinking and understanding.& In this video, a conversation takes place between Andrew Wang and Vitaliy Katsenelson. They discuss how writing for
two hours every morning benefits Katsenelson as both an investor and a portfolio manager by forcing them to identify broken links in their
thinking and gain a deeper understanding of concepts. & In this video, the speaker discusses the benefits of writing for two hours
every morning. They explain that writing helps them become a better
investor and portfolio manager by forcing them to identify broken
links in their thinking and gain a deeper understanding of concepts. \\
\textbf{Ratings:} N/A & \textbf{Ratings:}4 &\textbf{Ratings:}4 \\
\hline
\end{tabular}
}

\end{table*}
\begin{table*}[h]
\caption{Qualitative comparison
between best-performing baseline  with the proposed \textit{FASTER} model.}\label{qualitative}
\centering
\scalebox{0.7}{
\begin{tabular}{p{0.20\textwidth}|p{0.45\textwidth}|p{0.40\textwidth}|p{0.25\textwidth}}
\hline
\textbf{Gemma2-9b-Summary} & \textbf{MAF-TAV Summary} & \textbf{Generated Summary by \textit{FASTER}} & \textbf{Gold Summary} \\
\hline
Andrew and Mariko discuss financial advisory and their own personal experience on the topic. & The video discusses the topic of money and its various associations and emotions. Speaker 1 asserts that money is not mysterious and that there are various mindsets surrounding it, such as the extreme views of feral capitalism and the notion that money is evil. Speaker 1 believes that money is just a tangible representation of an energetic exchange and that it is essential to examine our relationship with it. & In this video, Andrew Wang and Mariko Gordon discuss the complex nature of money and the different mindsets surrounding it. Mariko emphasizes that money is just a tangible representation of an energetic exchange, and our emotions towards it can greatly influence its impact on us. \textbf{Financial planning will be effective. }& The discussion touches on the idea that handling money is not just about investing but involves many layers and dimensions... The conversation emphasizes examining our relationship with money and how it affects us. \\
\textbf{Ratings:} 2 & \textbf{Ratings:}3 &\textbf{Ratings:}4 &\textbf{Ratings:}5\\
\hline
\end{tabular}}
\end{table*}

\begin{table*}[!t]
\caption{Error Analysis of  \textit{FASTER}- model. Here the bold highlighted section depicts unnecessary generated phrases. }\label{errorA}
\centering
\scalebox{0.72}{
\begin{tabular}{p{0.46\textwidth}|p{0.37\textwidth}|p{0.40\textwidth}}
\hline
\textbf{Original transcript} & \textbf{Predicted Summary} & \textbf{Gold Summary} \\
\hline
Let's talk about earning and money matters!We know There's always something new to discover! Clear financial goals drive budgeting and investments. Whether for a house, retirement, or education, defined objectives ensure resource allocation. Historically, investing in Nifty 50 stocks, spanning diverse Indian sectors, yields consistent growth at 10-12\% annually, offering significant returns when aligned with specific goals. & Establishing clear financial goals \textbf{helps budgeting and long-term investment strategies saving for a house loan payment, retirement, or education expenses, health insurance}, Clear goals ensure proper allocation. Nifty 50 stocks historically yield 10-12\% growth, offering significant returns aligned with goals.
 & Whether saving for a house, retirement, or education, defined objectives ensure proper resource allocation budget planning is needed. Investing in Nifty 50 stocks, representing diverse sectors of the Indian economy, historically yields consistent growth at 10-12\% annually, highlighting the potential for significant returns aligned with specific financial goals.\\

\hline
\end{tabular}
}

\label{error}
\end{table*}

\begin{table}[]
\caption{Generalizibilty of \textit{\textit{FASTER}} model in different Financial NLP Tasks}\label{finNlp}
\centering
\scalebox{0.65}{
\begin{tabular}{l|l|l|llll}
\hline
{Datasets}& {Tasks}   & {Model Name} & \multicolumn{4}{l}{Evaluation Measurements}  \\ \cline{4-7} 
    & &    & \multicolumn{1}{l|}{F1-Score} & \multicolumn{1}{l|}{ROUGE} & \multicolumn{1}{l|}{BLEU}  & BERT Score \\ \hline
{F-sentiment} & {\begin{tabular}[c]{@{}l@{}}Sentiment\\ Classification\end{tabular}}  & Baseline& \multicolumn{1}{l|}{73.43}    & \multicolumn{1}{l|}{-}& \multicolumn{1}{l|}{-}& -\\ \cline{3-7} 
    & & Proposed& \multicolumn{1}{l|}{79.68}    & \multicolumn{1}{l|}{-}& \multicolumn{1}{l|}{-}& -\\ \hline
{X-Fincorp}   & {\begin{tabular}[c]{@{}l@{}}Complaint \\ Classification\end{tabular}} & Baseline& \multicolumn{1}{l|}{91.6}& \multicolumn{1}{l|}{-}& \multicolumn{1}{l|}{-}& -\\ \cline{3-7} 
    & & Proposed& \multicolumn{1}{l|}{87.51}    & \multicolumn{1}{l|}{-}& \multicolumn{1}{l|}{-}& -\\ \hline
{EDT-SUMM}    & {\begin{tabular}[c]{@{}l@{}}Unimodal \\ Summarisation\end{tabular}}   & Baseline& \multicolumn{1}{l|}{-}   & \multicolumn{1}{l|}{52.10} & \multicolumn{1}{l|}{34.60} & 90.83 \\ \cline{3-7} 
    & & Proposed& \multicolumn{1}{l|}{-}   & \multicolumn{1}{l|}{50.34} & \multicolumn{1}{l|}{34.28} & 87.78 \\ \hline
{MM-SUMM}& {\begin{tabular}[c]{@{}l@{}}Video \\ Summarisation\end{tabular}} & Baseline& \multicolumn{1}{l|}{-}   & \multicolumn{1}{l|}{32.41} & \multicolumn{1}{l|}{18.56} & 89.10 \\ \cline{3-7} 
    & & Proposed& \multicolumn{1}{l|}{-}   & \multicolumn{1}{l|}{30.23} & \multicolumn{1}{l|}{16.12} & 86.27 \\ \hline
\end{tabular}}
\end{table}
This section outlines the experimental setup, summarizes baseline approaches, compares results, and evaluates the efficacy of our \textit{FASTER} model. We assess the quality of the results through qualitative analysis and also scrutinize any limitations identified in the performance of our model. Furthermore, our research outcomes aim to tackle the research concerns mentioned below.\\
\textbf{Research Question 1 (RQ1):} Is \textit{FASTER} strong enough to surpass the performance of the baselines?\\
\textbf{Research Question 2 (RQ2):} Describe the impact of utilizing multiple modalities.\\
\textbf{Research Question 3 (RQ3):} To what extent does the proposed \textit{FASTER} model generalize to other tasks?\\
\textbf{Research Question 4 (RQ4):} What limits the model?\\
\textbf{Additional Implementation Details}\\The model,was fine-tuned with a learning rate of 2e-05 over 5 epochs on NVIDIA-A100 with 80GB. It utilized a batch size of 8 with gradient accumulation set to 2, employing the Adam optimizer with betas of (0.9, 0.999) and epsilon of 1e-08. A linear learning rate scheduler was applied, and mixed precision training using Native AMP was used. The model's performance was evaluated with a loss of 2.1986 on the evaluation set with a (70-20-10)\% training, eval and testing split. The evaluation compared performance against popular baselines which were fine-tuned on \textit{\textit{Fin-APT}} corpus: Long T5 \cite{raffel2020exploring}, BART \cite{lewis2019bart}, MAF-TAV \cite{kumar2022did}, GPT 3.5 \cite{openai2023gpt35}, FlanT5 \cite{chung2024scaling}, VideoLLaVA2-7b \cite{lin2023video}, VideoLLaMA2-7b \cite{damonlpsg2024videollama2} GPT-4o \cite{openai_gpt4} and LLaMA3-Instruct-1b \cite{dubey2024llama}, using ROUGE (R) \cite{lin2004rouge}, BLEU (B) \cite{papineni2002bleu}, and BERTSCORE \cite{zhang2019bertscore} metrics to assess the \textit{FASTER} framework's effectiveness in generating multimodal summaries. Moreover, SSIM, and RMSE effectively evaluate frame extraction quality by assessing structural and pixel-level accuracy. SSIM measures perceptual similarity in luminance, contrast, and structure, while RMSE calculates overall pixel-wise error.
\subsection{Analytical Discussion}
\begin{figure}[h]    
\centering
    \includegraphics[scale = 0.35]{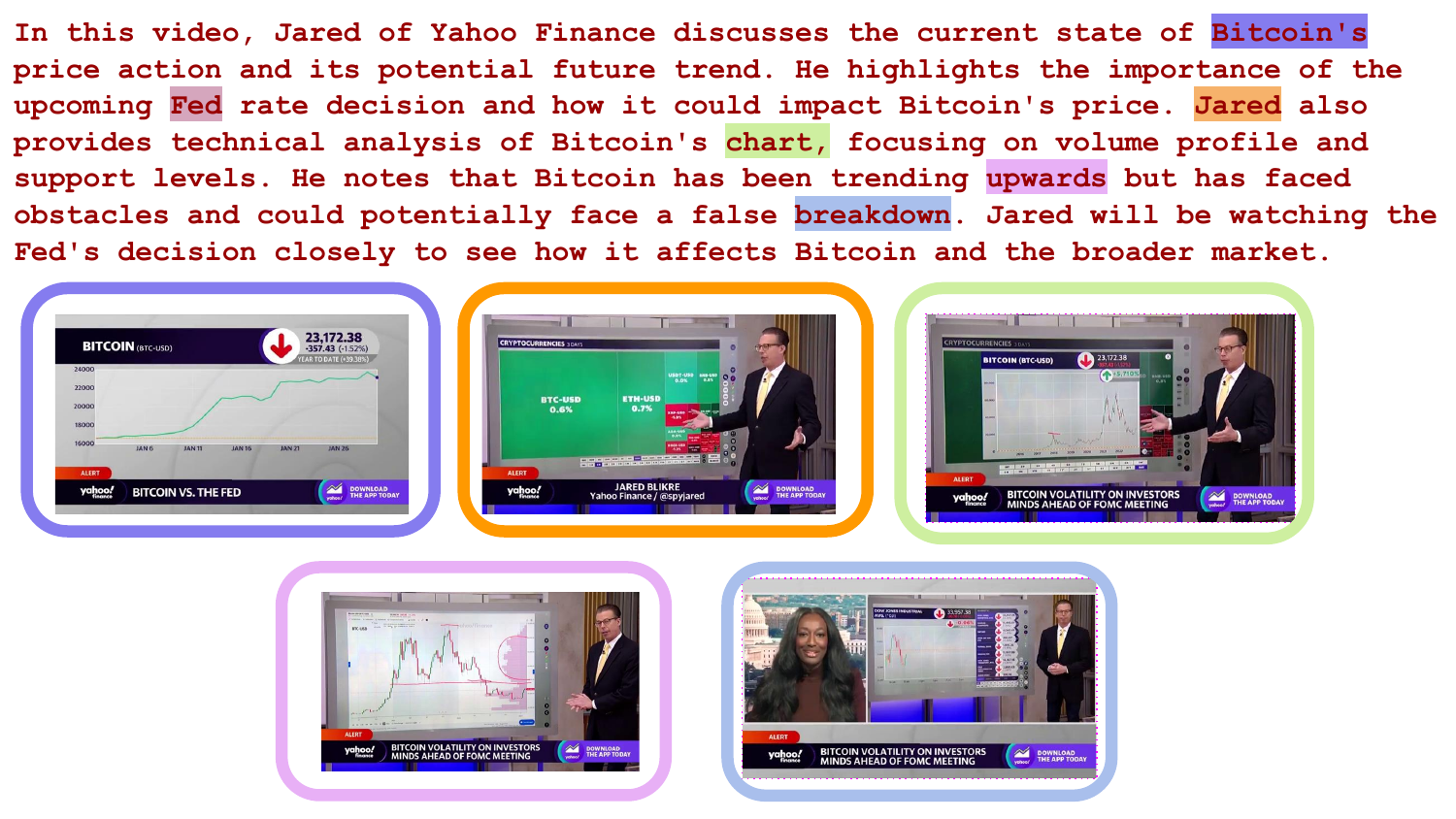}
    \caption{The summarized output of the \textit{FASTER} model features highlighted keyframes, with each frame corresponding to the highlighted colors in the advisory summary. }\label{FASTER-output}
    
\end{figure}

In this segment, we answered the potential research questions followed by qualitative and error analysis of the proposed model.
\textit{Note: Our primary goal is to generate concise financial summaries from extensive video advice, with a secondary aim of producing corresponding images aligned with key summary points. We explore these tasks in two configurations: one featuring BOS+DPO and one without (applicable only for LLMs, as VLM-based experiments were excluded due to their inherent visual features). The study evaluates the performance of the \textit{FASTER} model against other LLMs and VLMs for financial summary generation.}\\
\textit{\textbf{Answer to RQ1 \textit{FASTER'}s strength:}} As depicted in Table \ref{comp}, integrating BOS+DPO feature leads to higher performance for all the baseline models. Interestingly, \textit{FASTER} with BOS featured Gemma2-9b model performs significantly impressive and beats larger models such as GPT-4o, LLaMA3-Instruct-1b while adding modified DPO features, performance only leads to a higher score. Notably, a compelling comparative analysis emerges between large-scale, billion-parameter language models and their smaller counterparts. This observation is further substantiated, how incorporating BOS+DPO enhanced contextual embeddings and preference-aligned learning paradigms facilitates substantial improvements in model efficiency and scalability in summary generation tasks.\\
\textbf{\textit{Answer to research question 2 Modality Comparison:}} The utilization of multiple modalities significantly enhances model performance, as shown in Table \ref{Modalities}. While text-only input consistently outperforms visual-only input across all metrics, dynamic visual variations provide complementary contextual information, albeit with potential interference in textual feature extraction. Incorporating visual and speaker-related data (T+V+S) improves performance by approximately 5\% over text-only input, demonstrating the value of multimodality. Subsequently, Table \ref{Bos_comp} demonstrates that incorporating BOS features consistently enhances summary generation performance. The results indicate that the sequential integration of key elements from multiple modalities, such as speaker information, numerical data, graphs, and captions, significantly improves the model's ability to generate accurate, contextually rich, and informative multimodal summaries. Speaker segmentation and diarization allow the model to identify who is speaking and when, thereby ensuring accurate alignment of transcripts with visual context—even in cases involving off-screen speakers or multiple participants. As shown in Figure \ref{FASTER-output}, including the speaker’s name, such as "Jared," in the generated summary enhances factual precision and relevance. Overall, the inclusion of speaker information strengthens temporal alignment and semantic grounding, contributing to more accurate and context-aware multimodal summarization.

Additionally, to assess image quality as depicted in Figure \ref{FASTER-output}, we used a trainable scorer and the Naive/Cosine Similarity method. Evidently, in Table \ref{ranking}, the trainable scorer outperforms Naive/Cosine Similarity across key metrics: RMSE (0.412 vs 2.74), SSIM (0.44 vs 0.19), and F1 score (0.32 vs 0.09), indicating more accurate predictions, better structural similarity, and improved relevance identification. The Naive/Cosine Similarity method, relying on random frame selection, leads to inconsistent results, whereas the trainable scorer provides more reliable and precise outputs.
\noindent \textbf{\textit{Answer to RQ3 Generalizability:}}  Table \ref{finNlp} demonstrates the \textit{FASTER} model's adaptability across various financial NLP tasks, including sentiment and emotion detection \cite{dey2024socialite}, complaint classification, and single-stock prediction. Validation using the X-FinCORP dataset \cite{das2023let}, EDTSum, and SumScreen datasets (40-50 minute videos) \cite{papalampidi2022hierarchical3d} further highlights its robust performance and ability to generalize effectively across both unimodal and multimodal data. 

\noindent \textbf{\textit{Answer to RQ4 Limitations: }}The proposed modular approach is designed to capture comprehensive multimodal information from both video and audio, minimizing information loss that could compromise generation quality. However, BOS feature extraction introduces latency due to the time required for information aggregation. Additionally, the dataset predominantly features static backgrounds in advisory podcast videos, with limited visual elements such as stock prices, financial tables, and charts. Consequently, the model's reliance on the visual modality remains less significant compared to the textual modality.
\begin{figure*}[!hbt]    
\centering
    \includegraphics[scale = 0.42]{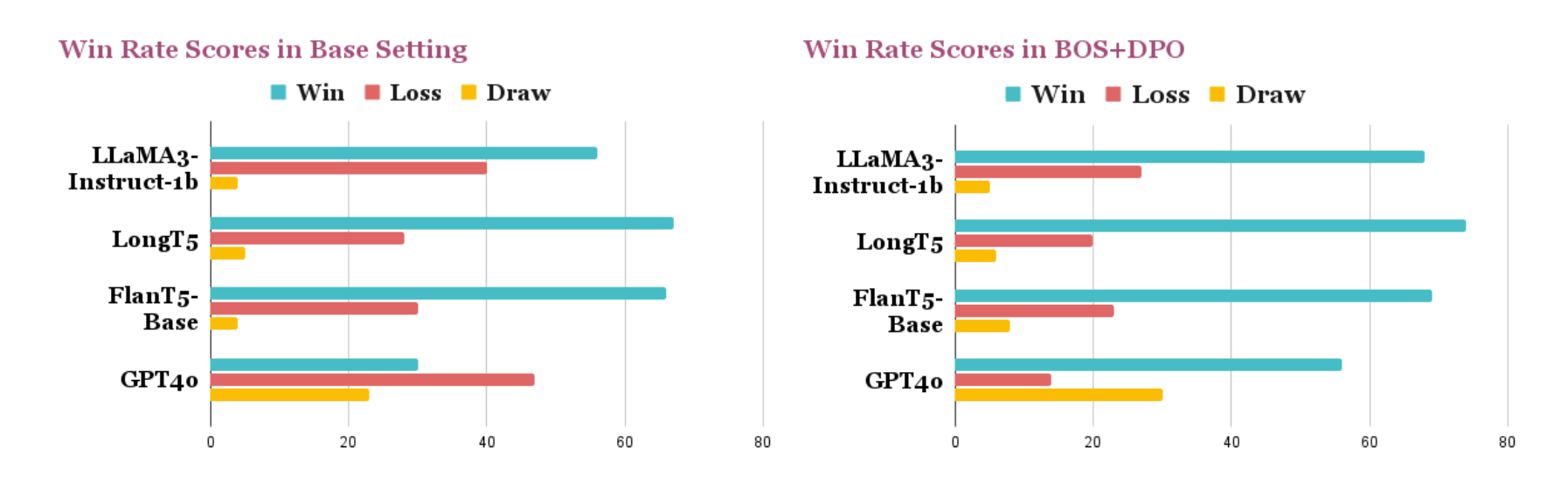}
    \caption{Expert evaluation of \textit{FASTER} model performance against the Baselines }\label{win}
    
\end{figure*}

\subsection{Comprehensive Scrutiny}

\textbf{Human Evaluation with Expert Scrutiny:}
We enlisted five financial experts from \textit{Category Alpha}, all of whom are active Spotify users accustomed to regularly listening to daily financial episodes lasting 30-40 minutes. To mitigate the potential limitations of automated metric assessments, we conducted a human evaluation involving 47 test samples. Figure \ref{win} demonstrates the visual representation of the influential performance of \textit{FASTER} with BOS+DPO featured Gemma2-9b model. 

Our model consistently demonstrates improved performance in winning outcomes compared to GPT-4o, while also reducing both drawing and losing rates. Although the base setting of the Gemma2-9b model performs well against other baseline models, it still falls short when compared to GPT-4o, which is a large billion-parameter model embedded with highly contextual information. This study highlights that the injection of BOS+DPO significantly enhances the scalability of smaller models.

\noindent To evaluate performance, we employed a tie-discounted accuracy metric, which compares the majority vote of the winning category with each expert's rating for every example. A score of 1 is given for a match, 0.5 for a tie, and 0 for a mismatch. Initially, we compared the performance of various baselines with our \textit{FASTER} model on 47 video samples, which was configured using the BOS+DPO-enhanced Gemma2-9b model. Subsequently, we recorded how often each baseline outperformed the proposed model and assigned E-FAIR ratings to the winning summaries. Category Alpha experts then rated the baseline summaries on a scale of "good" (2–3), "better" (3–4), or "best" (4–5), with the baseline receiving the highest cumulative ratings being declared the winner. Based on these evaluation scores, we calculated the expert-expert inter-agreement score to be 61.35\%. The evaluation focused on context coherence within the financial domain and was rated on a scale from 1 (inferior) to 5 (excellent). 

\noindent \textbf{Qualitative Analysis:} Table \ref{qualitative} presents key quality differences between Gemma2-9b, MAF-TAV and \textit{FASTER} with BOS+DPO featured Gemma2-9b. It shows how propsed model accurately generated a tangible representation of money and how its influence impacts emotions while maintaining summary quality. This demonstrates \textit{FASTER} model with BOS configuration's superiority with a 4 rating by offering a comprehensive and accurate summary with vital details. However, Gemma2-9b added irrelevant information, and MAV-TAV lacked relevant information. Thus, their generated summaries received an average rating of 3.5 and 2, respectively.\\
\textbf{Error Analysis:} The findings presented in Table \ref{errorA} undeniably showcase our model's proficiency in extracting the most relevant information from the aligned topic and preserving pertinent data. Nevertheless, since the language models are pre-trained on a large corpus, it is imperative to acknowledge that the model sporadically generates unnecessary extraneous information in the given context. For instance, phrases like \textit{'helps budgeting, health insurance'} are not required to comprehend the essence of the generated summary. 
\section{Conclusion}
This paper presents a distillation of concise insights from extensive financial videos for needy individuals. We introduce \textit{FASTER}, a modular approach with a modified DPO mechanism to present a condensed summary with aligned image keyframes using a ranker-based algorithm. We also present \textit{\textit{Fin-APT}}, a curated dataset of 470 multimodal financial advisory pep-talk videos validated by multiple financial experts. Our research shows that visual features contribute minimally compared to textual features in task-specific contexts and improve scalability even in the constrained data resource environment. Additionally, we integrated BOS features (BLIP-2, OCR, and Speaker Information) as facts to ensure summary coherence. The qualitative analysis demonstrates \textit{FASTER}'s superior summary generation, validating our approach. Conclusively, our research introduces a new chapter for multimodal summary generation, significantly contributing to the research community. 

{\bf Ethical Disclaimer: The article includes unsolicited feedback, inherent to the work. Prudent investment decisions require evaluating personal finances, risk tolerance, and goals. This feedback does not reflect the authors' views. The dataset is intended for academic research only and is not for commercial use. No video samples were reproduced in violation of content rights; the dataset was created from publicly available YouTube content without compensation or modification. Copyright remains with YouTube.

}

\section*{Acknowledgement}
This work is a joint collaboration between the Indian Institute of Technology Patna and CRISIL Data Science Limited. We extend our sincere gratitude to Tannishtha, and Jeel Doshi for their invaluable contributions as dataset annotators.

\bibliographystyle{ACM-Reference-Format}
\bibliography{reference.bib}










\end{document}